\documentclass[runningheads]{llncs}

\usepackage{eccv}

\usepackage{eccvabbrv}

\usepackage{graphicx}
\usepackage{booktabs}

\usepackage[accsupp]{axessibility}  

\usepackage{algorithm}
\usepackage{algpseudocode}
\usepackage{xcolor}
\usepackage[table, dvipsnames]{xcolor}
\usepackage{hhline}
\usepackage{array}
\usepackage{multirow}
\usepackage{enumitem}
\usepackage{wasysym}

\usepackage{pifont}

\newcolumntype{P}[1]{>{\centering\arraybackslash}p{#1}}
\newcolumntype{M}[1]{>{\centering\arraybackslash}m{#1}}
\newcommand{\PreserveBackslash}[1]{\let\temp=\\#1\let\\=\temp}
\newcolumntype{C}[1]{>{\PreserveBackslash\centering}p{#1}}
\newcolumntype{R}[1]{>{\PreserveBackslash\raggedleft}p{#1}}
\newcolumntype{L}[1]{>{\PreserveBackslash\raggedright}p{#1}}

\usepackage{tikz}

\usepackage{hyperref}

\usepackage{orcidlink}

\begin{document}

\title{A Unified Revisit of Temperature in Classification-Based Knowledge Distillation} 

\titlerunning{A Unified Revisit of Temperature in KD}

\author{Logan Frank \quad\quad\quad Jim Davis}

\authorrunning{L.~Frank and J.~Davis}

\institute{Ohio State University \\
\email{\{frank.580,\ davis.1719\}@osu.edu}}

\maketitle

\begin{abstract}
 A central idea of knowledge distillation is to expose relational structure embedded in the teacher's weights for the student to learn, which is often facilitated using a temperature parameter. Despite its widespread use, there remains limited understanding on how to select an appropriate temperature value, or how this value depends on other training elements such as optimizer, teacher pretraining/finetuning, etc. In practice, temperature is commonly chosen via grid search or by adopting values from prior work, which can be time-consuming or may lead to suboptimal student performance when training setups differ. In this work, we posit that temperature is closely linked to these training components and present a unified study that systematically examines such interactions. From analyzing these cross-connections, we identify and present common situations that have a pronounced impact on temperature selection, providing valuable guidance for practitioners employing knowledge distillation in their work.
\end{abstract}
\section{Introduction} \label{sec:introduction}

Knowledge distillation (KD) \cite{Hinton2015a} is an effective technique for compressing the information stored in a large pretrained ``teacher'' into a smaller, more efficient ``student''. In KD, the student is trained to match the soft-target outputs (or internal features) of the teacher when given similar inputs, which has been shown to improve the student's performance over training on the task directly. The core idea being that the student is learning the relational knowledge between classes that could only be successfully captured by a higher-capacity teacher.

To facilitate this transfer, temperature scaling \cite{Guo2017a} is often employed to control the output softness. In the original KD formulation \cite{Hinton2015a}, temperature is applied before the softmax as a fixed value shared by the teacher and student, followed by a KL-divergence loss. Since then, several works have either expanded on this paradigm \cite{Su2025a, Li2023b, Jin2023a} or argued that it is insufficient and proposed new ways of utilizing temperature (through new loss functions, etc.) \cite{Sun2024a, Huang2022a, Zhao2022a, Sun2025a}.

While there has been substantial progress and advancements made in KD, the answers to some fundamental questions remain unclear: \textit{``How should I apply temperature? What value should I use?''} Historically, we have observed cycles of newly proposed KD methods claiming to be better than shared fixed temperature followed by works showing that the original approach still works well \cite{Shen2020a, Hao2023a, Beyer2022a} (like eggs being good or bad for you depending on the year \cite{Myers2023a}). To this day, many companies employ output matching and divergence losses to distill their models \cite{Deshmukh2025a, Google2025a, Bai2025a}. Furthermore, finding the best temperature is often the result of an exhaustive grid search using one training configuration, without any sound reason as to why that value was the right choice (besides having the best score).

This discrepancy surrounding temperature leads one to feel uncertain on what to do. Moreover, other works often possess shortcomings which only exacerbate this uncertainty: 1) they only consider \textit{one} temperature value and set of hyperparameters, but in practice we know optimal choices are not always consistent across configurations, 2) their teacher models are \textit{only} trained from scratch, but the real-world often employs fine-tuned teachers, 3) they all utilize the same \textit{excessively small} (\texttt{<}2M parameters) students, but students can range up to 20M parameters in practice, and 4) they only consider well-curated, coarse-grained datasets whereas the real-world often involves fine-grained and/or few-example datasets. Works that do consider multiple training configurations, stronger teachers, or other datasets often only investigate that one component in isolation or do not elaborate on the cross-connections between the components \cite{Beyer2022a, Huang2022a, Qian2025a}.

In this work, we revisit temperature and its usage in KD through a unified study that systematically examines the interplay between temperature and other common KD training elements such as optimizer, teacher pretraining/finetuning, dataset granularity, and more. From extensive experimentation, we identify practical situations where specific temperature value ranges are favored over others, providing valuable insights that can reduce the need for an exhaustive and time-consuming grid search. Our contributions are summarized as follows:
\begin{enumerate}[noitemsep,nolistsep]
    \item We provide novel insights that identify specific KD scenarios which have a pronounced and consistent effect on the range of optimal temperature values.
    \item We show that in a common real-world KD situation, surprisingly large temperature values produce the best scores across many networks and datasets.
    \item We present a set of experimentally-grounded recommendations for future work/research in KD to promote positive and meaningful steps forward.
\end{enumerate}

\noindent We begin with a review of related work in Sect.~\ref{sec:related_work}. The various components of our study are described in Sect.~\ref{sec:methodology}. Lastly, extensive experiments and analysis of our findings are presented in Sect.~\ref{sec:experiments}.

\section{Related Work} \label{sec:related_work}

Knowledge transfer from large to small models was introduced in \cite{Bucilua2006a} and later formalized as ``knowledge distillation'' in \cite{Hinton2015a}. Subsequent work has examined the properties which influence the success of KD \cite{Urban2017a, Cho2019a, Beyer2022a, Hao2023a, Hao2025a, Frank2025a, Wu2024a, Hamidi2024a} and others have proposed structural changes to the seminal approach \cite{Romero2015a, Shen2020a, Tian2020a, Tarvainen2017a, Sun2024a, Jin2023a, Li2023b, Zhao2022a, Sun2025a, Huang2022a, Qian2025a, Wang2024a}. Works that introduce new methods can be divided into two categories: \textit{output matching} and \textit{feature matching}. We focus on the former as it is the most widely applicable strategy and used throughout industry, and therefore highlight those works. In \cite{Sun2024a, Sun2025a, Jin2023a, Li2023b, Su2025a}, the original divergence loss remains intact with new pre- and/or post-processing operations being introduced: multiple temperature values \cite{Jin2023a}, adversarial temperature adjustments \cite{Li2023b}, logit standardization \cite{Sun2024a}, and entropy-based reweighting \cite{Su2025a}. New loss functions were proposed in \cite{Zhao2022a, Sun2025a, Huang2022a, Qian2025a} to either emphasize non-target class matching \cite{Zhao2022a} or other correlation measures \cite{Sun2025a, Huang2022a, Qian2025a}. However, as noted in Sect.~\ref{sec:introduction}, these works have shortcomings that make it difficult to assess meaningful improvements, which is further emphasized by their limited adoption in practice \cite{Deshmukh2025a, Google2025a, Bai2025a}.

Many works have addressed some of these deficiencies by utilizing extensive training configurations/durations \cite{Beyer2022a, Hao2025a, Fan2024a}, finetuned teachers \cite{Beyer2022a, Huang2022a, Shen2020a, Hao2025a, Fan2024a}, more capable students \cite{Beyer2022a, Shen2020a, Hao2025a, Xu2024a}, and fine-grained datasets \cite{Beyer2022a, Shen2020a, Qian2025a, Xu2025a, Fan2024a}. For example, \cite{Beyer2022a} argue successful KD is a matter of ``function matching'' and ``patient distillation'', facilitated by MixUp \cite{Zhang2018a} and longer training periods, respectively. Teachers finetuned from large-scale pretrained weights are considered in \cite{Beyer2022a, Shen2020a, Huang2022a}, though \cite{Huang2022a} only considers one teacher finetuned on ImageNet1K whereas \cite{Beyer2022a, Shen2020a, Hao2025a} finetune on several different datasets. Larger students and/or different weight initializations of the student are considered in \cite{Beyer2022a, Shen2020a, Xu2024a}, however they present differing outcomes with respect to student weight initialization. 

Despite these advances, several gaps remain in our understanding. Multiple different training durations are considered in \cite{Beyer2022a}, but other variables such as optimization hyperparameters and temperature (\ie, cross-connections) are left as a grid search, emphasizing the best-performing configuration without systematically analyzing interactions among them. Moreover, the role of fine-grained datasets in \cite{Beyer2022a, Shen2020a, Xu2025a, Fan2024a} are treated as an experimental data point instead of being analyzed as a factor that could impact distillation. Finally, temperature has predominantly been an argmax-selected grid search value or studied through sensitivity analyses rather than a central object of investigation. Our work instead places temperature at the forefront and studies how it interacts with core components of the KD pipeline.


\section{Experimental Setup} \label{sec:methodology}

In this section, we describe the experimental setup of our study. More specifically: 1) our baseline setup and 2) the relevant cross-connections that will be expanded on and investigated with respect to temperature in our investigation (\eg, training parameters, teacher pretraining/finetuning, etc.).

\subsection{Baseline}

We begin by establishing our baseline setup that will be consistent across all major experiments. This baseline consists of two datasets, two teacher models, and two student models, described in their respective sections below.
\smallskip

\noindent \textbf{Datasets.} We employ two common image classification datasets, Pets \cite{Parkhi2012a} and CIFAR100 \cite{Krizhevsky2009a}, which are widely utilized in previous KD works \cite{Beyer2022a, Frank2025a, Qian2025a, Sun2024a, Sun2025a} and consist of 37 fine-grained classes and 100 coarse-grained, respectively. We consider additional datasets in one of our studies, which will be described later.
\smallskip

\noindent \textbf{Teachers and Students.} We consider two teacher architectures, ResNet50 \cite{He2016a} and ViT-S \cite{Dosovitskiy2020a}, along with two common edge-capable student architectures, ResNet18 \cite{He2016a} and MobileNetV4-Conv-Medium (MNv4) \cite{Qin2024a}. In each experiment, we will examine all combinations of teacher and student (for both datasets) to ensure our results generalize across models. Similarly, we add more teachers and students in relevant experiments, as will be described below.
\smallskip

\noindent \textbf{Output Supervision.} All experiments follow the standard output matching approach for KD \cite{Hinton2015a}, as it is the most general technique and still widely used (and potentially most commonly used) in practice for its simplicity and aforementioned generality \cite{Deshmukh2025a, Google2025a, Bai2025a}. Given a pretrained teacher $\mathcal{F}_T$ and a student $\mathcal{F}_S$, an example is passed forward through both networks to produce teacher and student logits $z_T$ and $z_S$, respectively. A loss is computed on these logits as $\mathcal{L}(z_T,\ z_S) = \omega(f(\sigma(z_T),\ \sigma(z_S)))$ where $\sigma$ is some pre-processing operation on the logits, $f$ is a specific loss equation, and $\omega$ is some post-processing operation. In the original formulation, $\sigma$ is the temperature-scaled softmax, $f$ is the KL-divergence, and $\omega$ is the identity function. Thus, $\mathcal{L}(z_T,\ z_S)$ becomes
\begin{align}
    \mathcal{L}(z_T,\ z_S) = {KL}(p_T\ ||\ p_S) = \sum_{i\ \in\ \mathcal{C}} \left[\ p_{T(i)} log\ p_{T(i)}\ \text{-}\ p_{T(i)} log\ p_{S(i)}\ \right]
\end{align}

\noindent where $\mathcal{C}$ is the number of classes and $p_{T} = \text{softmax}(z_T / \tau)$ with temperature $\tau$, and similar for $p_S$. For \textit{all} experiments, we only consider teacher output matching like \cite{Beyer2022a} and thus exclude the cross-entropy loss to ground-truth. As part of our study, we will examine other KD approaches that alter $\sigma$, $f$, and $\omega$.
\smallskip

\noindent \textbf{Training Details.} By default, we finetune all teachers from ImageNet1K-pre- trained weights made available by the \texttt{timm} library \cite{Wightman2019a, Steiner2022a, Wightman2021a}. The Pets and CIFAR100 teachers are finetuned for 20 and 40 epochs, respectively, using a label smoothed ($\alpha=0.1$) cross entropy loss. For CNN teachers, we use an initial learning rate of 5e-4 and a weight decay value of $0.05$ and for ViT, we use 5e-5 and $0.1$ for the learning rate and weight decay, respectively.

As for a baseline KD configuration, we use the AdamW optimizer \cite{Loshchilov2018a} with an initial learning rate of $0.002$ (decayed using a half-period cosine scheduler), a weight decay value of $0.05$, and a batch size of $64$. Distillation is performed for $1024$ and $512$ epochs for Pets and CIFAR100, respectively, using the original KD loss formulation (as just defined). We note that many of these decisions will be justified in experiments. Per-layer gradient clipping with an $L2$ norm threshold of $1.0$ is employed to stabilize training. Finally, we utilize data augmentation by first performing inception-style cropping \cite{Szegedy2015a} (for Pets only), RandAugment ($n=2$, $m=14$) \cite{Cubuk2020a}, and random horizontal flipping followed by a 50/50 chance of either MixUp \cite{Zhang2018a} or CutMix \cite{Yun2019a}, both with $\alpha\sim\mathcal{U}(0,1)$. We found this augmentation scheme to consistently perform the best in preliminary experiments.
\smallskip

\noindent \textbf{Temperature.} With the focus of this work being temperature, we employ several different values including some that, to the best of our knowledge, have not been explored previously. These values are $\tau\in\{1, 2, 3, 4, 5, 7, 10, 20, 40\}$, where a score will be provided for \textit{every} temperature in \textit{every} experiment.
\smallskip

\subsection{Dimensions of Interplay} \label{sec:interplay}
With our baseline experimental setup established, we now describe the different axes at which we will expand upon to investigate specific interactions with temperature. Many different ``dimensions'' of variation exist within neural network training. However, examining all possible combinations of choices is simply infeasible. Thus, we select dimensions which we believe could have a strong impact on temperature. This includes the \textbf{1)} KD approach, \textbf{2)} training configuration, \textbf{3)} teacher ``origination'', \textbf{4)} student initialization, and \textbf{5)} dataset granularity. Within each dimension, we augment our baseline setup to compliment the respective study. For example, utilizing four teachers in teacher origination, four datasets in dataset granularity, etc., with all other components being fixed.
\smallskip

\noindent \textbf{KD Approach.} Over the years, several different \textit{output matching} KD techniques have been proposed. However, as mentioned previously, we still see the original approach used heavily in practice. Thus, we compare some representative works with the original shared fixed temperature technique at different temperature values. We employ the following methods for this study: Decoupled KD \cite{Zhao2022a}, Entropy Adaptive KD \cite{Su2025a}, Global Curriculum \cite{Li2023b}, Logit Standardization \cite{Sun2024a}, Multi-Level KD \cite{Jin2023a}, Refined Logits \cite{Sun2025a}, and Stronger Teacher \cite{Huang2022a}. To implement these approaches, we use the code available in their GitHub repos then select hyperparameter values according to their respective papers. 
\smallskip

\noindent \textbf{Training Configuration.} There are many decisions that can be made in the context of optimization/training. Assuming an optimal learning rate and weight decay are selected, some of the most important choices remaining include optimizer, batch size, and number of training epochs. In this study, we expand across those dimensions by examining \textit{all possible combinations} of two optimizers (AdamW and SGD), two batch sizes (64 and 256), and five values of training epochs ($\{2^6, ..., 2^{10}\}$ for Pets and $\{2^5, ..., 2^{9}\}$ for CIFAR100) to analyze how temperature behaves with each of these different configurations. 
\smallskip

\noindent \textbf{Teacher Origination.} We define the ``origination'' of a model as the process taken to reach its current state, this includes general pretraining (\eg, ImageNet1K/21K) and/or target training (training from scratch or finetuning). Here, we examine two common teacher originations: 1) target trained from random weights and 2) ImageNet pretrained then target finetuned. Moreover, we consider finetuning for four different durations (\ie, number of epochs) to evaluate the importance of general pretraining knowledge. Lastly, we employ two additional teachers in the forms of ConvNeXt-T \cite{Liu2022a} and RegNetY-6.4GF \cite{Radosavovic2020a}. The primary goal of this study is to investigate how the teacher origination impacts temperature in KD, but we will analyze architecture effects as well.
\smallskip 

\noindent \textbf{Student Initialization.} As there are many ways to obtain a teacher, there are similarly many ways to initialize a student, which has been explored to varying degrees \cite{Beyer2022a, Shen2020a}. Here, we investigate the interplay between student initialization and temperature by examining the following initialization strategies: 1) random, 2) target trained from random, 3) general pretrained, and 4) general pretrained then target finetuned weights. Note these weights serve as the starting point for the student in KD. We additionally add two students, ConvNext-P \cite{Woo2023a} and MobileViTv2-1.5 \cite{Mehta2023a}, to this investigation and like teacher origin, we will analyze architecture effects as a byproduct of our study.
\smallskip

\noindent \textbf{Dataset Granularity.} The final piece of our study is to examine the interaction between temperature and coarse-grained vs. fine-grained dataset classes. One fine-grained dataset (Pets) and one coarse-grained dataset (CIFAR100) are already considered in the previous studies, thus we examine two other datasets to accompany those results. We employ Cars \cite{Krause2013a} and Tiny ImageNet \cite{Deng2009a} for the new fine-grained and coarse-grained datasets, respectively. Experiments in this study will be similar to those from the training configuration inquiry, but now with these new datasets to further understand the effects of dataset granularity.
\smallskip


\section{Experimental Results} \label{sec:experiments}

We present the results of our KD temperature studies following the same order as Sect.~\ref{sec:interplay}. We reiterate that \textit{all} experiments are conducted using \textit{all} possible teacher-student combinations for \textit{both} datasets, and furthermore for \textit{all} temperature values. Note that each accuracy score is the mean result from three seeds. 
\smallskip 

\refstepcounter{subsection} \label{subsec:approach}
\noindent \textbf{4.1\quad KD Approach.} We began our study by first investigating the question: \textit{``How should I apply temperature?''} In other words, when we introduce finetuned teachers, modernized training regimes, and more capable students (all of which are not present in many previous image classification KD works \cite{Sun2024a, Sun2025a, Zhao2022a}), does any one KD approach stand out amongst the others? Thus, we performed KD using our baseline setup and compared the original KL-divergence with shared fixed temperature technique to the other KD methods mentioned in Sect.~\ref{sec:interplay}.

\begin{figure}[t]
    \centering
    \includegraphics[width=\linewidth]{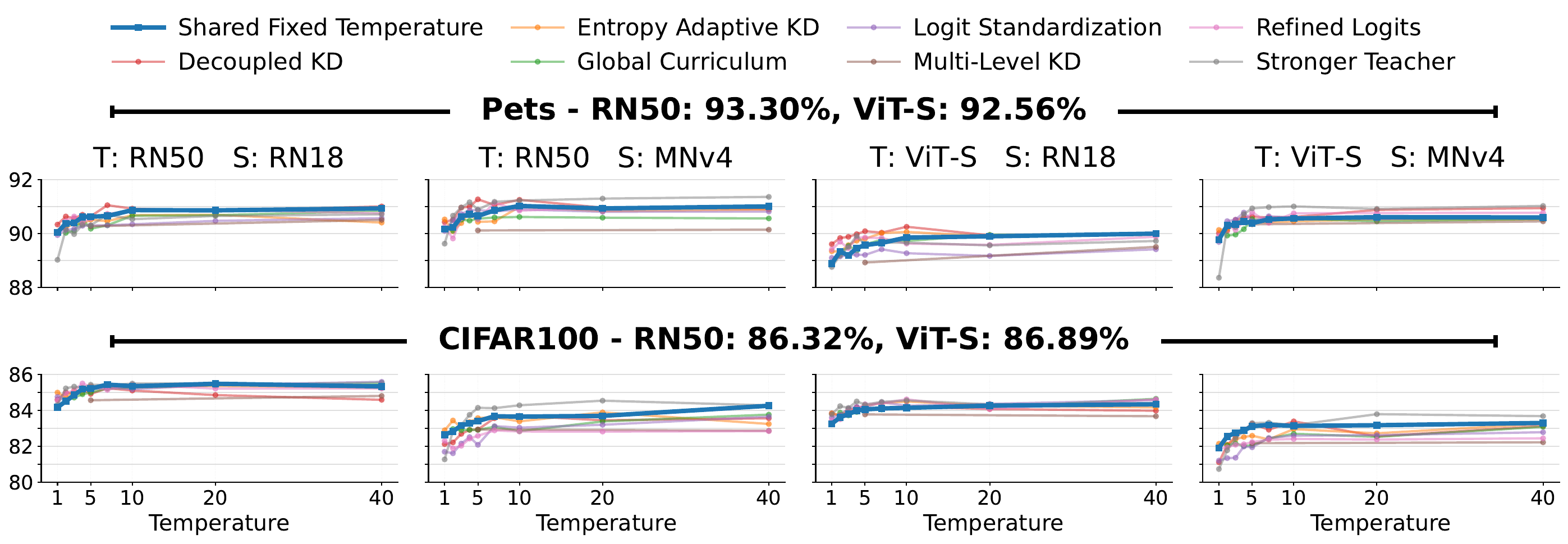}
    \caption{Accuracy of students distilled using various KD methods and temperatures.}
    \label{fig:methods}
\end{figure}

As shown in Fig.~\ref{fig:methods}, there does not appear to be one single approach that consistently beats all other techniques which, coupled with the simplicity of shared fixed temperature, might explain why the original strategy is still heavily used in practice. Given these results, we utilize shared fixed temperature with a KL-divergence loss as our mode of distillation for the remaining experiments. We do note that we examined non-shared and/or non-fixed temperature values in initial experiments, but did not observe any significant differences from the shared fixed version that warranted the extra complexity.

\smallskip

\refstepcounter{subsection} \label{subsec:config}
\noindent \textbf{4.2\quad Training Configuration.} Now that we have established shared fixed temperature to still be a viable approach for KD, we next considered possible alterations in the training configuration to begin to answer the question: \textit{``What value should I use?''}. As mentioned in Sect.~\ref{sec:interplay}, there are several ways that a training configuration could be modified. Assuming optimal (or near-optimal) learning rate and weight decay values are selected, we elected to investigate the influence of optimizer, batch size, and number of distillation epochs with respect to temperature. We acknowledge that other configuration choices exist, such as the usage of sharpness-aware minimization \cite{Foret2021a}, stochastic depth \cite{Huang2016a}, and more, but such tools are not utilized universally whereas optimizer, batch size, and training duration are obviously required. Thus, we distilled each of our teacher-student-dataset combinations for five different durations with all possible combinations of AdamW and SGD optimizers and 64 and 256 batch sizes.

\begin{figure}[t]
    \centering
    \includegraphics[width=\linewidth]{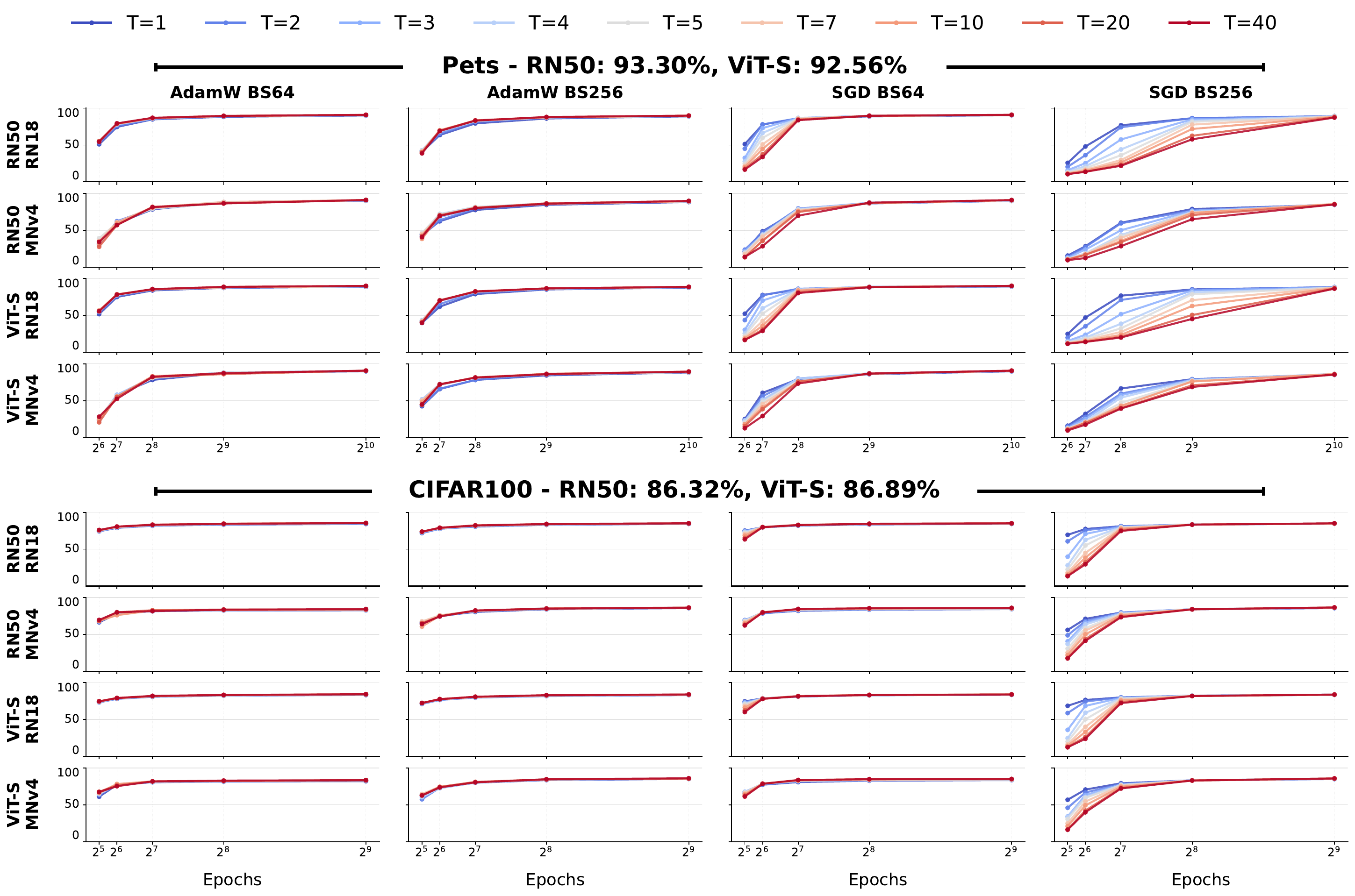}
    \caption{Accuracy of students distilled with different temperatures for various durations (number of epochs) using different combinations of optimizer and batch size.}
    \label{fig:config_all}
\end{figure}

\begin{figure}[t]
    \centering
    \includegraphics[width=\linewidth]{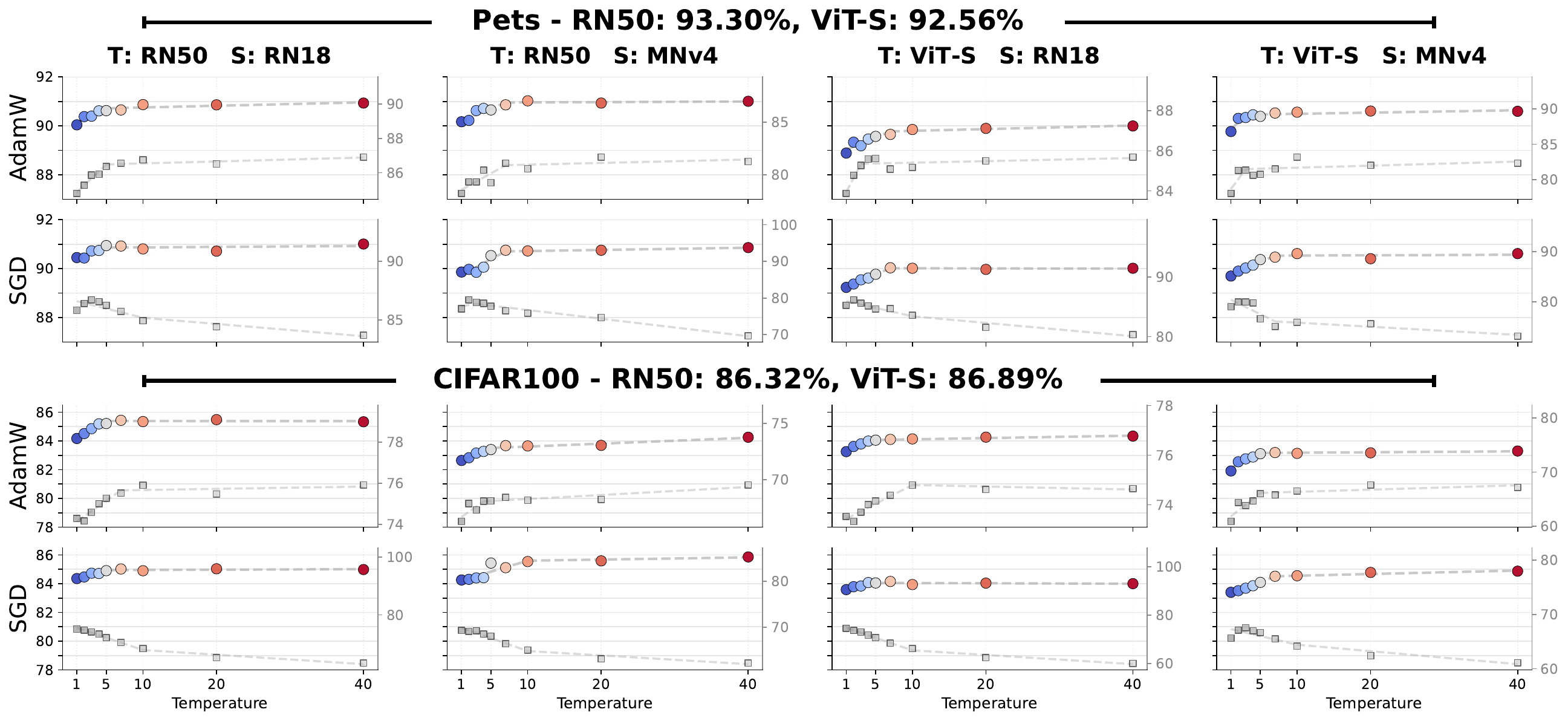}
    \caption{Accuracy of students at various temperatures. Color and grayscale denote evaluation at largest number of epochs and an earlier point (using right y-scale), respectively.}
    \label{fig:config_final}
\end{figure}

We see in Fig.~\ref{fig:config_all} that the optimizer, batch size, and number of distillation epochs all have significant interplay with temperature, producing large differences in the final outcomes of the student models. Between AdamW and SGD, we observe that AdamW is much more robust to the value of temperature. With SGD, smaller temperature values tend to perform substantially better with fewer epochs, but with longer training durations larger temperature values outperform. This ``cross over'' is emphasized in the SGD rows of Fig.~\ref{fig:config_final} where we plot the accuracy scores for the students trained with the longest duration (colored points) and overlay the accuracy scores for students trained for less epochs (grayscale points). Moreover, we observe that with an increased batch size, the point at which larger temperature values become better shifts to where a student needs more training to reach that cross over. 

\begin{figure}[t]
    \centering
    \includegraphics[width=\linewidth]{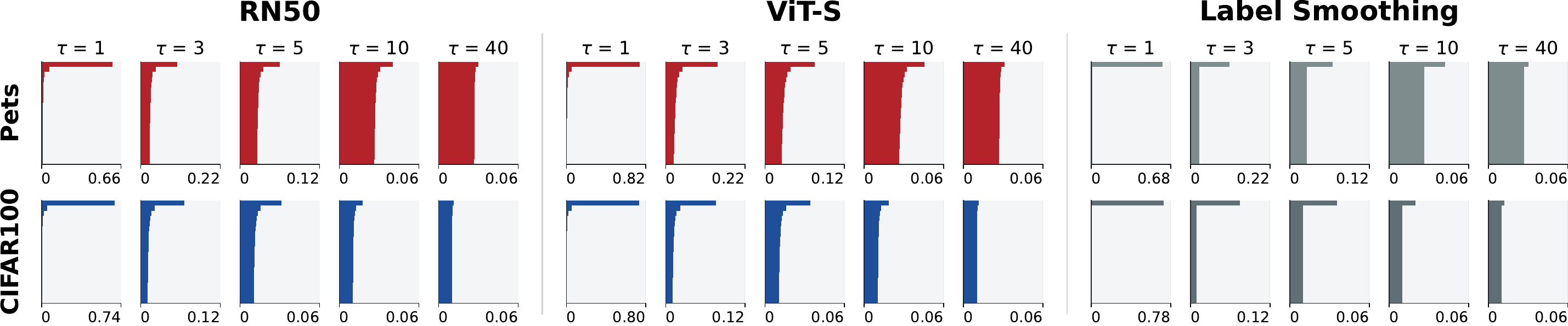}
    \caption{Colored figures: average sorted temperature-scaled softmax distributions of training samples passed through the teachers (truncated to top-20 classes). Grayscale figures: label smoothing distributions with entropies similar to the softmax distributions. Note the x-axis scales reducing with increasing temperature.}
    \label{fig:softmax_distributions}
\end{figure}

From analyzing the actual points in Fig.~\ref{fig:config_final}, we also notice that \textit{unexpectedly large} temperature values ($\tau\geq10$) surprisingly work well. There is a recognizable positive trend between temperature and accuracy from $\tau=1$ to $10$, with insignificant diffrences afterwards for most figures. We find this particularly interesting since, up to this point, most works have only considered temperatures between 1 and 5 with very few considering a value up to 10. However, with values as large as 40 one must wonder about the appearance of the teacher's temperature-scaled softmax outputs. Thus, we computed the average sorted softmax distributions output from the teachers and present those distributions in Fig.~\ref{fig:softmax_distributions}. At $\tau=10$, we see a <0.01 difference between the argmax class and all other classes with <0.0001 difference among the bottom classes. If we changed the x-extents to be $[0,1]$, the softmax outputs would appear essentially uniform. 

\begin{figure}[t]
    \centering
    \includegraphics[width=\linewidth]{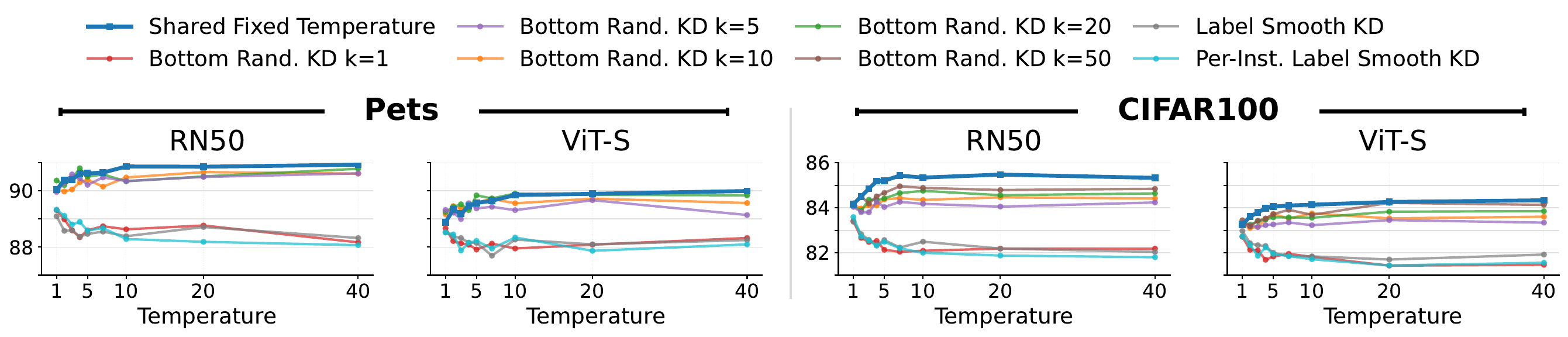}
    \caption{Accuracy of ResNet18 students distilled using various methods that alter the amount of relationship information provided by the teacher during training.}
    \label{fig:relationships}
\end{figure}

When these differences in softmax scores are so miniscule, one could then question the importance of class relationships. To analyze this, we conducted an experiment with two different supervisory signals: label smoothing and bottom $N$-$k$ randomized. For label smoothing, we created two variants: 1) \textit{global}, which selects a single label smoothing coefficient $\alpha$ for all examples such that the entropy of that distribution is similar to that of the average softmax distribution and 2) \textit{per-instance}, which now computes a cofficient for every example such that the entropy matches that sample's temperature-scaled softmax distribution. With both variants, all non-max classes now have the exact same score. The bottom $N$-$k$ randomized method preserves the top-$k$ softmax scores for an $N$ class distribution and randomizes the indexes of the remaining $N$-$k$ scores, thus exposing different amounts of class relationship information. The results of this experiment are shown in Fig.~\ref{fig:relationships} where we see that, even when the softmax distributions are nearly uniform at larger temperatures, the minor ($\pm$0.0001) per-class differences from standard temperature-scaling do still provide rich class relationship information for the student.
\smallskip

\refstepcounter{subsection} \label{subsec:origination}
\noindent \textbf{4.3\quad Teacher Origination.} As defined in Sect.~\ref{sec:interplay}, the teacher's ``origination'' is the means at which it reached its current state (\eg, large-scale general pretraining into target finetuning or direct target training from random weights). As there are many paths a model could take before becoming a teacher for KD, one must ask: \textit{``Should I tune temperature to my teacher?''} To explore this, we finetuned four different teacher models (our two baseline networks plus ConvNeXt-T and RegNetY-6.4GF) for four different durations and trained the three CNN teachers from scratch then distilled those models to our baseline students. We did not include a ViT-S teacher trained from random initialization as ViTs are notoriously known to only perform well when large-scale general pretrained \cite{Liu2021a}.

In addition to test accuracy, we also report two different relative entropy scores. Relative entropy is computed as $\hat{H}(p) = H(p) / H(\mathcal{U}_{\mathcal{C}})$ where $H(\cdot)$ is the entropy function, $p$ is a softmax distribution, and $\mathcal{U}_{\mathcal{C}}$ is the discrete uniform distribution for $\mathcal{C}$ classes. The two variants we consider are $\hat{H}(p)$, which utilizes the full softmax distribution, and $\hat{H}(\bar{p})$, which utilizes the softmax distribution \textit{without} the argmax score (normalized to sum to 1 before computing the entropy). 

Looking at Table \ref{tab:teacher_origination}, we observe two distinct trends as we increase the number of teacher finetuning epochs (red heatmap): 1) the student's performance degrades and 2) smaller temperature values begin to perform as well as larger ones. The decrease in performance can be explained by our entropy measures. Across all models, as we increase the number of finetuning epochs the value of $\hat{H}(p)$ \textit{decreases} toward 0 (\ie, one-hot), but the value of $\hat{H}(\bar{p})$ \textit{increases} toward 1 (\ie, uniform). Essentially, the teacher's softmax outputs are transforming into one-hot distributions that resemble label smoothed distributions when temperature-scaled, which we just showed in Fig.~\ref{fig:relationships} leads to performance degradation. Regarding smaller temperatures beginning to perform better, this can be justified in two ways. First, we additionally saw in Fig.~\ref{fig:relationships} that label smoothing preferred a temperature of $\tau=1$, so if the teacher outputs are becoming more like label smoothed distributions, then it would make sense that smaller temperatures would start to perform similar (or better) than larger ones. This is then validated when we look at the results of the teachers trained from random initialization (blue heatmap), where a temperature of $\tau=1$ often performs best. Therefore, it is likely that the excessively finetuned teacher is unlearning the general pretraining knowledge which supplied the rich class relationship information and in the case of the teacher trained from scratch, it simply did not learn any meaningful relationship information. Moreover, we see that the behaviors we have observed thus far persist across all teacher architectures (higher temperature being better with minimal finetuning, etc.). We also examined other general pretrained weights in additional experiments (\eg, different ImageNet1K, ImageNet21K, and CLIP \cite{Radford2021a}) and saw similar outcomes except in the case of the CLIP pretrained teachers. We leave further investigation as future work.

\begin{table}[t!]
    \centering
     \caption{Accuracy heatmap of students distilled from teachers with different originations using various temperatures. Red and blue heatmaps denote finetuned and randomly initialized teachers, respectively. Entropy $\hat{H}(p)$/$\hat{H}(\bar{p})$ is defined in Sect.~\ref{subsec:origination}.}
    \label{tab:teacher_origination}
    \begin{tabular}{c}
        \includegraphics[width=0.99\linewidth]{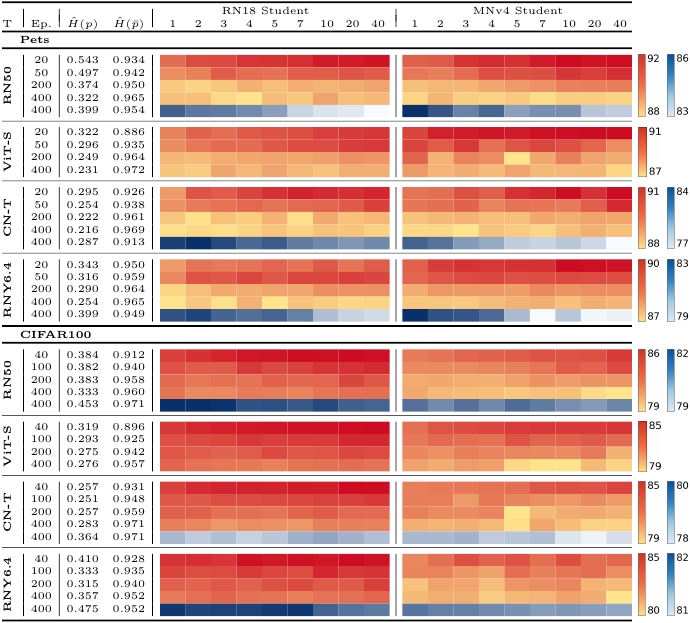}
    \end{tabular}
\end{table}

\smallskip

\refstepcounter{subsection} \label{subsec:initialization}
\noindent \textbf{4.4\quad Student Initialization.} We just showed the impact that general pretraining has on the teacher with respect to temperature, which has traditionally been something reserved for only the larger models (>20M parameters). However, we are now seeing some providers start to ImageNet1K-pretrain models with <10M parameters (\ie, potential candidates for students). Thus, if one were to consider these pretrained weights as the initialization for their student in KD: \textit{``Does temperature even matter with my initialization?''} We investigated this question by initializing four student models (our two baseline networks plus ConvNeXt-P and MobileViTv2-1.5; Sect.~\ref{sec:interplay}) using random (Rand), target trained (TT), general pretrained (IN1K), and general pretrained then target finetuned (FT) weights then performed KD using our baseline teachers. 

\begin{table}[t!]
    \centering
     \caption{Accuracy of differently initialized students distilled from baseline teachers using various temperatures. The best scores (and ties) are emphasized and underlined.}
    \label{tab:student_initialization}
    \begin{tabular}{c}
        \includegraphics[width=0.99\linewidth]{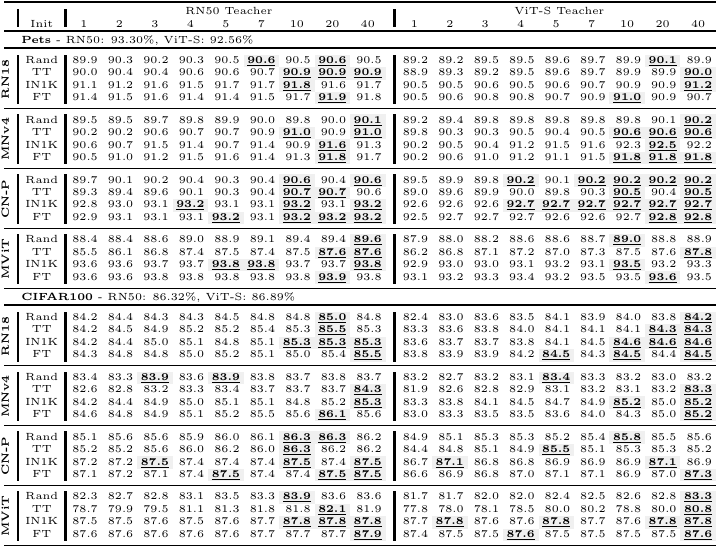}
    \end{tabular}
\end{table}

As shown in Table \ref{tab:student_initialization}, we continue to see the same benefit of employing larger temperatures across all student initializations and architectures, albeit a smaller gain for the general pretrained and target finetuned initializations. However, with the inclusion of the general pretraining then target finetuning initialization, the benefits of KD come into question (\ie, does KD provide any additional gains on top of target finetuning). After finetuning, the ResNet18, MobileNetV4, ConvNeXt-P, and MobileViTv2 students achieved test accuracy scores of 89.72\%, 90.98\%, 92.80\%, and 93.51\%, respectively, for the Pets dataset and 82.31\%, 81.68\%, 85.63\%, and 86.07\%, respectively, for CIFAR100. Thus, we can conclude that KD does still provide performance gains that are not obtainable from training with ground truth labels only. As all of our students are $\sim$10M parameters, we conducted further experiments with $\sim$3M parameter students and saw our findings extend to these more compact models as well. 


\smallskip

\refstepcounter{subsection} \label{subsec:dataset}
\noindent \textbf{4.5\quad Dataset Granularity.} In all previous experiments, we considered Pets and CIFAR100 datasets where Pets contains fine-grained classes and CIFAR100 contains more coarse-grained classes. To round out our study, we investigated two more datasets, Cars and Tiny ImageNet, to answer the question: \textit{``Should I pick a higher or lower temperature for my dataset?''} Adding these datasets now provides us with insights on two fine-grained (Pets and Cars) and two coarse-grained (CIFAR100 and Tiny ImageNet) datasets so we can better understand how class granularity affects our choice of temperature.

\begin{figure}[t!]
    \centering
    \includegraphics[width=\linewidth]{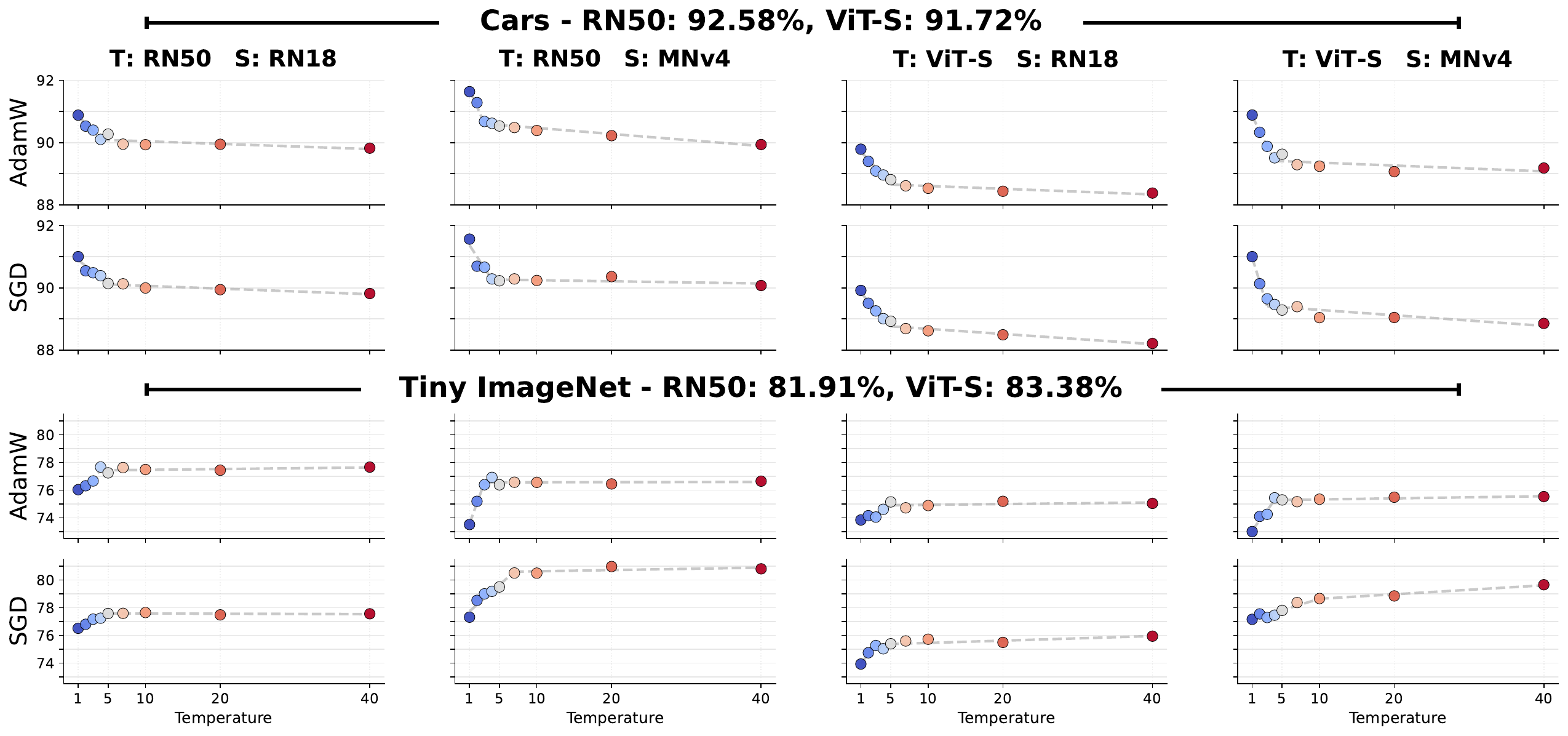}
    \caption{Accuracy of students at various temperatures distilled using additional datasets.}
    \label{fig:dataset_granularity}
\end{figure}

From Fig.~\ref{fig:dataset_granularity}, we notice the same positive trend in Tiny ImageNet that we saw with the Pets and CIFAR100 datasets. One observation we can make is that it appears as though the ``inflection point'' of temperature vs. test accuracy (\ie, the temperature value at which test accuracy begins to plateau) shifts closer to 1. This intuitively makes sense as each class in a coarse-grained dataset may only have 2-3 other classes that it shares relationships with whereas a fine-grained dataset could have 10 or more. Thus, a larger temperature is needed to expose the entire relationship hierarchy for fine-grained datasets. However, when we look at Cars, we notice a \textit{negative} trend contrary to what was observed previously. 

\begin{figure}[t]
    \centering
    \includegraphics[width=\linewidth]{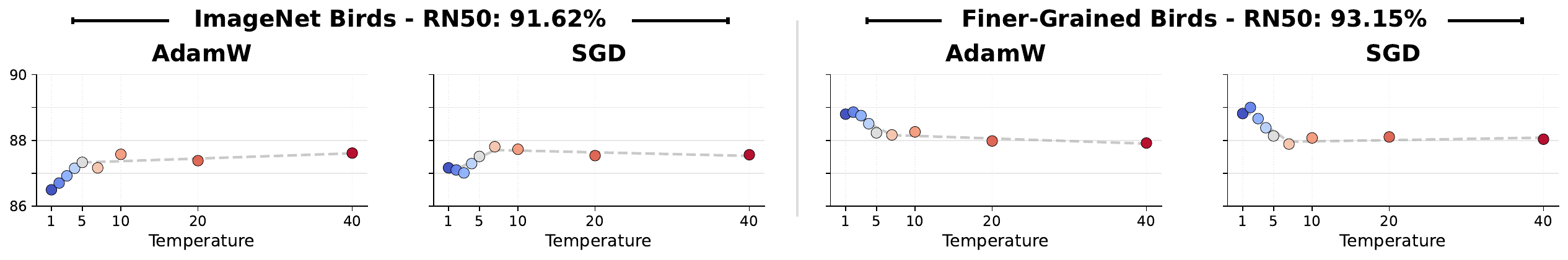}
    \caption{Accuracy of ResNet18 students at various temperatures distilled using a ResNet50 teacher and fine-grained datasets with different levels of class granularity.}
    \label{fig:dataset_granularity_birds}
\end{figure}

We investigated this phenomenom further by creating two new datasets: ImageNet Birds and Finer-Grained Birds. Our hypothesis is that the teacher does not have a strong understanding of the fine-grained inter-class relationships in Cars since the general pretraining dataset (ImageNet1K) only contains generic vehicle classes such as ``passenger car'' and ``pickup''. Thus, we created the ImageNet-Birds dataset by collecting \textit{new} images of the same ImageNet1K classes under the ``bird.n.01'' node in the WordNet hierarchy \cite{Deng2009a}. Since some of the bird classes in ImageNet1K are generic birds (\eg, ``magpie'', ``vulture''), we can expand these classes into finer-grained versions to create the Finer-Grained Birds dataset (simulating the Cars dataset). For this dataset, we collected images of ``black-billed magpie'', ``green magpie'', ``black vulture'', ``king vulture'', etc. from existing bird classification datasets \cite{Wah2011a, Birds525a}. For fair comparison, both datasets have 52 classes, 30 training images per class, and 20 test images per class. 

As shown in Fig.~\ref{fig:dataset_granularity_birds}, the positive trend between temperature and test accuracy exists for the ImageNet-Birds dataset, but the negative trend seen with the Cars dataset is also demonstrated by the Finer-Grained Birds dataset. It should be noted that most (if not all) classes in Pets, CIFAR100, and Tiny ImageNet exist in ImageNet1K. Thus, we can resolve that if the finetuning dataset does not contain exact or near-exact class matches to the general pretraining dataset, then it is likely that smaller temperature values will be preferred in KD.

\section{Discussion} \label{sec:discussion}

This work revisited the usage of temperature in KD by first answering the question \textit{``how should I apply temperature?''} in which we ``rediscovered'' that the original shared fixed temperature approach \cite{Hinton2015a} still works well in classification-based KD (Sect.~\ref{subsec:approach}). We then proceeded to analyze the question \textit{``what value should I use?''}, which was investigated through a unified study that systematically examined the interactions between temperature and \textbf{1)} training configuration, \textbf{2)} teacher origination, \textbf{3)} student initialization, and \textbf{4)} dataset granularity. Through analyzing these temperature-centric cross-connections, we observed several interesting behaviors that emerged from our results:

\begin{itemize}[noitemsep,nolistsep]
    \item Adaptive optimizers (AdamW) exhibit greater robustness to temperature. 
    \item With SGD, smaller temperature values tend to perform better with less training, but larger values eventually outperform them. 
    \item If sufficiently long training is achieved then unexpectedly large temperature values can work well in KD, but is only enabled when the teacher has a solid understanding of the dataset's inter-class relationships which we found via 1) general pretraining and minimal target finetuning and 2) a finetuning dataset that shares substantial class overlap with the pretraining dataset. 
    \item When a larger temperature value ($\tau\geq10$) is employed, the class relationships still matter, even when the softmax differences are minor ($\pm$0.0001). 
    \item Fine-grained datasets (that overlap classes with the pretraining dataset) tend to prefer larger temperatures than coarse-grained ones, likely due to needing a higher value to expose the full relational structure of the data. 
\end{itemize}

\noindent In addition, we observed other notable properties not specific to temperature: 1) SGD and/or larger batch sizes encourage ``patient distillation'' \cite{Beyer2022a}, 2) longer finetuning decreases KD efficacy, and 3) general pretrained or pretrained then target finetuned student initializations perform exceptionally well. To the best of our knowledge, this collective set of ideas has not been previously expressed in literature, with previous works largely utilizing temperature values \texttt{<}5 \cite{Sun2025a, Su2025a, Zhao2022a} and others only examining certain components in isolation \cite{Beyer2022a, Huang2022a, Hao2025a, Frank2025a}.



Lastly, based on our experimental results, we propose some recommendations for future research: 1) evaluate methods on trained-from-scratch \textit{and} finetuned teachers, 2) demonstrate multiple temperature values and/or other forms of interaction, 3) examine additional datasets (\eg, finegrained ones), and 4) employ other training regimes (longer training, mixup/cutmix, etc.). Beyond KD, we also encourage model developers to continue pretraining their smaller networks. 

\section{Conclusion} \label{sec:conclusion}

We performed a systematic study to revisit temperature in knowledge distillation and to investigate its interplay with other important KD components. Through our study, we conducted experiments and analyses that give insight into how specific details in 1) training configuration, 2) teacher origination, 3) student initialization, and 4) dataset granularity can have a drastic impact on the student under various temperature values. Most notably, we showed that smaller temperatures can perform better earlier in training, with unexpectedly large values performing better with longer training, particularly for minimally finetuned teachers Additionally, we demonstrated that fine-grained datasets tend to prefer greater values compared to coarse-grained datasets. Building on our findings, we outlined a set of recommendations to help guide future work in KD.

\newpage
\section*{Acknowledgements}

We thank the DoD HPCMP for the use of their computational resources. We would also like to thank Matthew Scherreik, Elizabeth Sudkamp, and Vincent Velten from the U.S. Air Force Research Laboratory for their assistance.

%
%
\bibliographystyle{splncs04}
\bibliography{main}
\end{document}